\documentclass[conference]{IEEEtran}
\IEEEoverridecommandlockouts
\usepackage{cite}
\usepackage{amsmath,amssymb,amsfonts}
\usepackage{algorithmic}
\usepackage{graphicx}
\usepackage{textcomp}
\usepackage{xcolor}
\usepackage{url} % 7.17.24 - Needed to fix weird spacing in references with urls
\usepackage{subcaption} % Added 7.22.24 for combined figure
 \usepackage{hyperref}

\def\BibTeX{{\rm B\kern-.05em{\sc i\kern-.025em b}\kern-.0for storingT\kern-.1667em\lower.7ex\hbox{E}\kern-.125emX}}
\begin{document}
\bstctlcite{IEEEexample:BSTcontrol} % Added 7/17/24, jake

% \title{Graph Integration for Diffusion-Based Manifold Alignment\\
% \thanks{This material was supported by a grant from the Simmons Research Endowment.}

% \author{\IEEEauthorblockN{Jake S. Rhodes\textsuperscript{$\dagger$}}
% \IEEEauthorblockA{\textit{Department of Statistics} \\
% \textit{Brigham Young University}\\
% Provo, Utah, USA \\
% rhodes@stat.byu.edu}
% \and
% \IEEEauthorblockN{Adam Rustad\textsuperscript{$\dagger$}}
% \IEEEauthorblockA{\textit{Department of Computer Science} \\
% \textit{Brigham Young University}\\
% Provo, Utah, USA \\
% arusty@byu.edu}
% }

\title{Graph Integration for Diffusion-Based Manifold Alignment
 \thanks{\textsuperscript{$\dagger$}Both authors contributed equally to this work. This material was supported by a grant from the Simmons Research Endowment.}}

\author{
    \IEEEauthorblockN{Jake S. Rhodes\textsuperscript{$\dagger$}}%
    \IEEEauthorblockA{
        \textit{Department of Statistics} \\
        \textit{Brigham Young University}\\
        Provo, Utah, USA \\
        rhodes@stat.byu.edu
    }
    \and
    \IEEEauthorblockN{Adam G. Rustad\textsuperscript{$\dagger$}}%
    \IEEEauthorblockA{
        \textit{Department of Computer Science} \\
        \textit{Brigham Young University}\\
        Provo, Utah, USA \\
        arusty@byu.edu
    }
}

% \thanks{\textsuperscript{$\dagger$}Both authors contributed equally to this work.}

\maketitle

\begin{abstract}
Data from individual observations can originate from various sources or modalities but are often intrinsically linked. Multimodal data integration can enrich information content compared to single-source data. Manifold alignment is a form of data integration that seeks a shared, underlying low-dimensional representation of multiple data sources that emphasizes similarities between alternative representations of the same entities. Semi-supervised manifold alignment relies on partially known correspondences between domains, either through shared features or through other known associations. In this paper, we introduce two semi-supervised manifold alignment methods. The first method, Shortest Paths on the Union of Domains (SPUD), forms a unified graph structure using known correspondences to establish graph edges. By learning inter-domain geodesic distances, SPUD creates a global, multi-domain structure. The second method, MASH (Manifold Alignment via Stochastic Hopping), learns local geometry within each domain and forms a joint diffusion operator using known correspondences to iteratively learn new inter-domain correspondences through a random-walk approach. Through the diffusion process, MASH forms a coupling matrix that links heterogeneous domains into a unified structure. We compare SPUD and MASH with existing semi-supervised manifold alignment methods and show that they outperform competing methods in aligning true correspondences and cross-domain classification. In addition, we show how these methods can be applied to transfer label information between domains.
\end{abstract}


\section{Introduction}

Data collected from an individual observation can be derived from various sources, modalities, or times, but can be intrinsically linked, such as brain scans from multiple technologies~\cite{boehm2022multimodalintegration}, or documents translated into various languages~\cite{koehn2005europarl}. In many cases, information derived from multiple sources can be more rich than from a single source~\cite{meng2020survey-data-fusion}. In such applications, large quantities of high-dimensional data are collected and analyzed. However, the intrinsic structure of the data can often be captured by an underlying lower-dimensional representation~\cite{izenman2012introduction}. One common representation of these structures is a manifold---a topological space that, at a local scale, resembles a Euclidean space.

Manifold alignment is a set of methodologies that attempt to find a shared underlying manifold representation of the data sources. The learned representation or embedding can be subsequently used to explore inter-domain relationships or to improve the knowledge gained through subsequent machine learning tasks, such as label prediction.

Finding a shared representation that incorporates multiple types of data can be challenging due to the disparate nature of the datasets and the extent to which known links or correspondences are known. For example, single-cell measurements cannot be directly linked to epigenetic measurements collected through bulk sequencing, but some shared information can be found in both domains~\cite{welch2017ma-singlecell, amodio2018magan}, although no one-to-one correspondence is possible. On the other hand, when comparing brain scans from different sources but from the same patient, correspondences between images from different sources can be inferred.

Manifold alignment seeks to identify a shared representation of distinct but related datasets while retaining the unique characteristics of each and emphasizing their commonalities. This representation should maintain some semblance of the structure that the original datasets possess, helping to preserve the inherent patterns and relationships within the data. Manifold alignment transforms each dataset into a lower-dimensional, shared latent space in which locally similar instances within each dataset---as well as corresponding instances across different datasets---are positioned close together~\cite{lafon2006datafusion, nguyen2020PracticalCM}. A key assumption in manifold alignment is that the intrinsic structures of the distinct modalities lie on the same underlying manifold~\cite{wang2011manifold}. The latent space representing this manifold should maintain geodesics along the manifold.

Manifold alignment methods are categorized according to the degree of known correspondence between observations of different domains. Cases without known correspondences fall under the umbrella of unsupervised manifold alignment---this is akin to unsupervised machine learning where label or auxiliary information is not used or present~\cite{wang2009ma-wo-correspondence, cui2014guma, stanley2020harmonic-alignment}. Supervised manifold alignment employs one-to-one correspondence of points~\cite{lindenbaum2020multi-view-dm} or label information (e.g., classes)~\cite{wang2011heterogeneous-da, tuia2016kema, duque2023mali}. Semi-supervised alignment falls on the middle ground, where partial correspondence is known. This can take the form of different views of known entities, such as various translations of the same source document~\cite{ham2005ssma, wang2008ma-procrustes}, or some features are common to more than one domain~\cite{amodio2018magan}.

In this paper, we introduce two semi-supervised manifold alignment methods inspired by two distinct manifold learning models, Isomap~\cite{tenenbaum2000isomap} and Diffusion Maps~\cite{coifman2006dm}. Isomap seeks to find a manifold representation of data by estimating geodesic distances via shortest paths across a locally connected graph structure. Our corresponding method, \textbf{S}hortest \textbf{P}aths on the \textbf{U}nion of \textbf{D}omains (SPUD) forms a unified graph structure by using known correspondences to form new graph edges through which inter-domain relationships can be learned. That is, SPUD learns new edges between points in co-domains $\mathcal{X}$ and $\mathcal{Y}$ in the form of inter-domain geodesic distances. The resulting graph structure serves as a precomputed kernel that can be used in subsequent manifold learning models.

Shortest path distances learn meaningful embeddings when the data is not inherently noisy, but are topologically unstable otherwise~\cite{balasubramanian2002isomap-stability}. To overcome this weakness, we present a second, diffusion-based approach that is more robust to noise and insensitive to data density. The method is called \textbf{M}anifold \textbf{A}lignment by \textbf{S}tochastic \textbf{H}opping (MASH). MASH learns a local geometry within each domain and forms a joint diffusion operator using known correspondences to link co-domains. New interdomain correspondences are learned through iterative random domain ``hopping''. By learning new edge weights, we form 1) a graphical structure that can be used in several manifold learning algorithms and 2) a coupling matrix that serves as a basis for cross-domain projection.

We compare SPUD and MASH with existing semi-supervised manifold alignment methods and show that our methods often outperform existing ones at aligning true correspondences and performing cross-domain classification. Additionally, we demonstrate the effectiveness of our methods at transferring label information between domains when one domain is sparsely labeled. This can have positive implications in use cases where labeling in one domain is expensive. 

\section{Related Works}

Although some of the following methods generalize to more than two modalities, we focus on manifold alignment in the context of two modalities or co-domains, denoted as $\mathcal{X}$ and $\mathcal{Y}$. We limit the comparative works to semi-supervised methods where some known correspondence exists between points in the different domains.

Ham et al.~\cite{ham2005ssma} introduced two methods for semi-supervised manifold alignment. In the first approach, some known coordinates in the embedding space are given. The embedding function from both domains is constrained to map corresponding points to these coordinates. The second method is more akin to the situations we describe in this paper. That is, they have partially annotated correspondences and seek to preserve these correspondences in low dimensions. To this end, the authors constrain the cost function for Laplacian Eigenmaps to minimize the difference between embedding coordinates for known correspondences. We call this method SSMA and compare the resulting alignments with SPUD and MASH.

Lafon et al.~\cite{lafon2006datafusion} describe a multimodal data fusion process for manifold alignment based on Diffusion Maps~\cite{coifman2006dm}. They begin by performing density normalization on similarity graphs, thus ensuring that the method is invariant to data density. After normalization, they apply Diffusion Maps to embed each domain separately into a lower-dimensional space, attempting to capture the intrinsic geometry within each domain. They learn an affine transformation to align the embeddings using the iterated closest point algorithm~\cite{fitzgibbon2003icp} and geometric hashing~\cite{wofson1997geom-hashing}. The authors do not provide open-source code or enough details to replicate their results. Thus, we do not include this method in our comparisons.

The authors of~\cite{wang2008ma-procrustes}, begin the process of manifold alignment by mapping each domain to lower-dimensional embeddings using Laplacian Eigenmaps~\cite{belkin2003laplacian}. Following the dimensionality reduction, the authors use a subset of known correspondences between the domains to perform Procrustes analysis~\cite{gower2004procrustes}, which aligns the embeddings and minimizes the differences between the corresponding points, thus attaining a common coordinate system. Their experimental results demonstrate that the matching probabilities achieved through this approach generally outperform those obtained using SSMA~\cite{ham2005ssma}. We call this approach Manifold Alignment via Procrustes Analysis or MAPA and include this method in our comparisons.

A similar approach to MAPA was done in~\cite{shen2017manifold-matching}, where the authors construct a set of ranked distances from which they determine the shortest paths within a given domain. They apply multidimensional scaling (MDS) using the set of shortest paths as a precomputed distance. To align the two MDS embeddings, they apply Procrustes matching. They compared their results with manifold learning algorithms Isomap, Locally Linear Embedding, and MDS at matching corresponding points in lower dimensions using the common Swiss Roll dataset. Their matching ratio shows the number of points correctly matched with their corresponding points in the lower-dimensional representation. However, the authors did not compare with other alignment methods.

In the paper~\cite{wang2011manifold}, the authors construct an embedding using a joint graph Laplacian. To create the joint Laplacian, similarity measures are first learned from the $k$-NN distances within each separate domain using a Gaussian kernel. The similarity matrices, which we denote as $W_{\mathcal{X}}$ and $W_{\mathcal{Y}}$, are used to form the diagonal block entries of the joint similarity matrix $W$, and represent the inter-domain similarity measures. The inter-domain similarity measures are defined to be non-zero only for points with known correspondence. Laplacian Eigenmaps are finally applied to the joint Laplacian to determine the joint embedding. We call this method Joint Laplacian Manifold Alignment, or JLMA and include it in our comparisons.

Manifold Alignment Generative Adversarial Network, MAGAN, was introduced in~\cite{amodio2018magan}. Two separate generator networks map from domain $\mathcal{X}$ to $\mathcal{Y}$ and vice versa, attempting to produce a correspondence in the other domain. This semi-supervised method aligns manifolds using three separate loss functions: a reconstructive loss, which compares a point in $\mathcal{X}$ to itself after mapping to $\mathcal{Y}$ and back again, a discriminator loss to be able to distinguish between a true and generated point, and a correspondence loss, which can be adapted for specific problems, but generally penalizes reconstructions that are too dissimilar from a known, true correspondence. The experiments demonstrated in the paper were limited to cases where each co-domain contained shared features, but one-to-one correspondence was not annotated. We adapt the MAGAN correspondence loss for our comparisons to handle cases where no features are shared between co-domains. Instead, we use known correspondences to ensure that alternative representations from each domain map to points similar to their alternative representation.

% The authors of~\cite{Kriebel2022uinmf} perform alignment using non-negative matrix factorization on a combined matrix composed of gene data from different sources. Like the displayed use cases in~\cite{amodio2018magan}, the matrix concatenates the datasets along shared features. Since the method requires shared features in each co-domain, we do not include this method in our results.

More recently, \cite{duque2023dta} introduced a method called Diffusion Transport Alignment (DTA) which combines diffusion methods with regulated optimal transport~\cite{villani2009optimaltransport, courty2014domain-reg-optimal}. The process starts by forming local similarity matrices in each domain by applying the $\alpha$-decaying kernel~\cite{moon2019phate} to $k$-NN graphs. Each matrix is row-normalized to form stochastic matrices (diffusion operators) $P_{\mathcal{X}}$ and $P_{\mathcal{Y}}$. As with traditional Diffusion Maps, these diffusion operators are powered to simulate random walks between points with each domain of $t$ discrete time steps. Subsets of these operators, representing transitions between points in co-domains with known correspondence, are multiplied to form cross-domain transition matrices used to join co-domains. The cosine distance is calculated between the rows of the cross-domain transition matrices and the powered diffusion operators. The result is an inter-domain distance matrix which is used in the regularized optimal transport problem. The authors showed that DTA outperformed MAGAN, MAPA, and SSMA for semi-supervised alignment. We include DTA in our comparisons.

The result of some methods, such as JLMA and SSMA, is an embedding of the manifold representation of the joint space. However, it can be advantageous for an alignment method to produce a mechanism to relate observations between domains directly. For example, MAGAN derives generative functions that act as projections between spaces. Thus, a data point can be directly mapped to the alternative domain via a trained generator. DTA forms a coupling matrix that can be used for cross-domain, barycentric projection~\cite{duque2023dta}. Similar projection matrices have been developed for unsupervised methods~\cite{demetci2022scot}. The joint diffusion operator of MASH can serve as a coupling in the form of cross-domain transition probabilities. The coupling matrix can assist in multi-domain analysis by estimating connection probabilities to determine a correspondence between points or estimate label transfer from one domain to another. 

\section{Graph-Based Manifold Learning}

A challenge of alignment is to find meaningful relationships between points across multiple data sets. The alignment process can be approached through dimensionality reduction with constraints induced by the known correspondences among the data points, ensuring that the reduction process respects the relationships between similar data points~\cite{wang2011manifold}. 

The embedding produced through manifold learning should accurately represent the intrinsic manifold structure, keeping the salient points intact by encoding meaningful representations of the local and global data structure. In this framework, graphs serve as powerful tools for storing the data structure and constraints linking modalities. Here we provide some details underlying graph-based manifold learning processes.

Many methods for manifold learning and alignment begin with learning a local data structure in the form of a symmetric, weighted graph ~\cite{tenenbaum2000isomap, coifman2006dm, duque2023dta, wang2011manifold, ham2005ssma}. Both Isomap and classical Diffusion Maps form their local similarity representations as a $k$-NN graph created by applying a kernel function---such as the Gaussian kernel $K(x_i, x_j; \sigma) = \exp\left(\frac{||x_i - x_j||^2}{\sigma}\right)$---to local Euclidean distances, forming non-negative weights that reflect the similarity between points $x_i$ and $x_j$. In~\cite{moon2019phate}, the authors developed the $\alpha$-decaying kernel, Equation~\ref{eq:alpha-decay}, to overcome the weaknesses of a fixed kernel bandwidth of the Gaussian kernel and improve connectivity between points from sparse data regions. The kernel is defined as

\begin{equation}
    \footnotesize
    \label{eq:alpha-decay}
    K_{k, \alpha}\left(x_i, x_j\right)=\frac{1}{2} \exp \left(-\frac{\left\|x_i-x_j\right\|^\alpha}{\sigma_k^\alpha\left(x_i\right)}\right)+\frac{1}{2} \exp \left(-\frac{\left\|x_i-x_j\right\|^\alpha}{\sigma_k^\alpha\left(x_j\right)}\right)
\end{equation}

where $\sigma_k(x_i)$ is the distance from $x_i$ to its $k$-th nearest neighbor and $\alpha > 0$ depicts the decay rate of the exponential. We use this kernel function to initialize our methodologies to learn graph edges.

From the graph, the global data structure can be learned. For example, Isomap~\cite{tenenbaum2000isomap} learns the global data structure by approximating geodesic distances by learning the shortest paths between graph nodes. Diffusion-based models, such as Diffusion Maps~\cite{coifman2006dm}, learn the global structure by simulating a random walk across the graph structure, thereby mapping all possible transitions from one point to another in $t$ discrete time steps. Through such processes, a continuous manifold is approximated using a discretized graph.

Building upon this premise, we introduce two methods for manifold alignment. In each case, local geometries are formed within domains $\mathcal{X}$ and $\mathcal{Y}$ separately. From these, we learn cross-domain global data structure, as outlined in Section~\ref{sec:methods}.

% Typically, manifold learning algorithms are applied to data from a single domain or modality. However, multiple data sources can be integrated to form a unified whole through manifold alignment. That is, data from multiple sources can be viewed as belonging to the same set so long as there is some intrinsic connection between them. This aligns with the view that data collected from the same individual should present common patterns regardless of measurement errors presented by different instrumentation***. 

% \subsection{Graph Learning}

\section{Methods}\label{sec:methods}

Within each domain, $\mathcal{X}$ and $\mathcal{Y}$, we define similarity matrices $W_{\mathcal{X}}$ and $W_{\mathcal{Y}}$ by applying the $\alpha$-decaying kernel (Equation~\ref{eq:alpha-decay}) to the distances of the $k$ nearest neighbors of each point. We note that the choice of distance metric is data-specific, but we use Euclidean distance in our experiments. Within each domain, the distances are 0-1 normalized.

Akin to~\cite{wang2011manifold}, we define a cross-similarity as a block matrix that relates points between and within domains. Defining $W_{\mathcal{X}\mathcal{Y}}$ to be the cross-domain similarity matrix, the typical assignment of similarity between the corresponding points $x_i$ and $y_j$ is determined to be:

\[
\begin{cases}
    W_{\mathcal{X}\mathcal{Y}}(i, j) = \nu & \text{if } x_i, y_j \text{ correspond} \\
    W_{\mathcal{X}\mathcal{Y}}(i, j) = 0 & \text{if } x_i, y_j \text{ do not correspond}
\end{cases}
\]

for $0 < \nu \le 1$. The magnitude of $\nu$ determines the strength of similarity between known correspondence.

We work under the assumptions that (1) the corresponding points serve as alternative representations of themselves in the co-domain, and (2) nearby points in one domain should remain close in the other domain, though differences in collected features and selection of distance metric will impact the type and degree of measured similarity. Under these assumptions, we justify defining the similarity between corresponding points to be maximal ($\nu = 1$), although in cases where these assumptions are deemed too strong, the user may choose to assign the inter-domain correspondence to $\nu < 1$. In the case of two domains, the self-similarity matrix is then defined to be

\[W = \left( \begin{array}{cc} W_{\mathcal{X}} & W_{\mathcal{X}\mathcal{Y}} \\ W_{\mathcal{Y}\mathcal{X}} & W_{\mathcal{Y}} \end{array} \right)\]
where $W_{\mathcal{Y}\mathcal{X}} = (W_{\mathcal{X}\mathcal{Y}})^{T}$.

In addition to assigning non-zero similarities between points of known correspondence, we can directly extend the influence of known correspondences to other points in the co-domain by accounting for the edge weights of the nearest neighbors. We extend $W_{\mathcal{X}\mathcal{Y}}$ and $W_{\mathcal{Y}\mathcal{X}}$ to include these influences. That is, we define $W_{\mathcal{X}\mathcal{Y}}(i, l) = \gamma W_\mathcal{Y}(l, j)$, $0 \le \gamma \le 1$, if (1) $x_i$ corresponds to $y_j$, (2) a correspondence to $y_l$ is unknown, and (3) $y_l$ is among the $k$ nearest-neighbors of $y_j$. The parameter $\gamma$ determines the influence of nearest neighbors in determining correspondence to neighboring points, which is typically set to 1.

A sufficient requirement to use this method of alignment initialization is the ability to define a measure of local similarity between points and a set of indices denoting the known correspondences. That is, we simply need to know how to measure similarity within a given domain (note that the measure of similarity can be different from one domain to the next), and how to designate similarities between domains via corresponding points. After the assignment of cross-domain similarities, several approaches can be used to learn an alignment; that is, to learn a way to map points to a shared embedding representation of a shared manifold. The cross-domain similarities, $W$, can be directly used in several manifold learning algorithms as a precomputed similarity matrix (or distance, e.g., 1 - $W$). This was done in JLMA~\cite{wang2011manifold}, which uses a similar representation for $W$ in Laplacian Eigenmaps~\cite{belkin2003laplacian}. However, the direct use of $W$ in an algorithm does not learn any additional correspondence between points of co-domains, unless edge learning is part of the manifold learning process. 

In addition to known correspondences, we seek to learn new connections, or graph edges, before mapping to low dimensions. In the following sections, we provide the details of learning edges and an alignment mechanism via shortest paths (SPUD) or cross-domain diffusion via a probabilistic random walk (MASH). 

% We apply two different methods to learn additional information. The first method is based on shortest paths or geodesics, more akin to the structure-learning mechanism of Isomap~\cite{tenenbaum2000isomap}. The second method learns both cross-domain local and global structures via the creation of new graph edges through a probabilistic random walk between domains via known correspondences. Additional details are provided in the subsequent sections.

\subsection{Shortest Paths on the Union of Domains (SPUD)}

We introduce some notation and definitions to facilitate the description of our methods. We denote the distance between points as $d(\cdot, \cdot)$, while the estimated geodesic distance within domain $\mathcal{X}$ is $d_{G_\mathcal{X}}(\cdot, \cdot)$, within domain $\mathcal{Y}$ as $d_{G_\mathcal{Y}}(\cdot, \cdot)$, and for the combined graph structure as $d_{G}(\cdot, \cdot)$. We call points with known correspondences anchor points. The set of all known anchors is denoted as $\mathcal{A}$, where $|\mathcal{A}| \le \min(|\mathcal{X}|, |\mathcal{Y}|)$. We denote the nearest anchor to point $x_i$ in domain $\mathcal{X}$ as $a_{x_i}$, that is, $d_{G_\mathcal{X}}(x_i, a_{x_i}) = \min\{d_{G_\mathcal{X}}(x_i, a_{x_k})| a_{x_k} \in \mathcal{A}\}$. The correspondence (or alternative representation) of $a_{x_i}$ in the co-domain $\mathcal{Y}$ is $\tilde{a}_{x_i}$. 

The graph structure learned using the $\alpha$-decaying kernel encodes the local relevance between points in a given domain. We form a union of the graphs of each domain by defining edges between known anchor points with edge weight $\nu$. The unified graph structure allows us to learn the relationships between points of different domains, forming the multimodal geodesic structure. To learn inter-domain relationships, we estimate the geodesic distance between all points in each domain via the known anchors. 

A correspondence between points $x_i$ to $y_j$ does not guarantee that $\tilde{a}_{x_i} = a_{y_j}$,  due to noise or difference in the type of distance measure. Supposing this correspondence is unknown, two ``shortest'' paths exist connecting $x_i$ to $y_j$. One path covers the shortest distance within $\mathcal{X}$: $x_i \rightarrow a_{x_i}, \quad \tilde{a}_{x_i} \rightarrow y_j$, while the other covers the shortest distance traversed in domain $\mathcal{Y}$: $y_j \rightarrow a_{y_j}, \quad \tilde{a}_{y_j} \rightarrow x_i$. To determine a single estimate of the geodesic distance, we can apply an aggregation function to them, such as the minimum, maximum, mean, or minimum absolute difference between the point-to-anchor distances.

% that is, we define 

% \[
% D_1 = |d_{G_{\mathcal{X}}}(x_i, a_{x_i}) - d_{G_{Y}}(\tilde{a}_{x_i}, y_j)|
% \] and

% \[
% D_2 = |d_{G_{\mathcal{Y}}}(y_j, a_{y_j}) - d_{G_{X}}(\tilde{a}_{y_j}, x_i)|
% \] then

% \[
% d_G(x_i, y_j) = \min(D1, D2)
% \]

 An alternative approach to form the alignment is to apply an information distance such as the potential distance~\cite{moon2019phate} or similar variants~\cite{duque2019dig} to the rows of the cross-domain geodesic distances. As a final step, the information distances are embedded via MDS~\cite{kruskal1964mds} to learn the joint manifold.

\subsection{Manifold Alignment via Stochastic Hopping (MASH)}

Excessive noise levels in the underlying manifold decrease the probability of learning the true geodesic distances. To overcome this potential weakness of SPUD, we introduce MASH as a more robust alternative to learn the global, inter-domain data structure via diffusion~\cite{coifman2006dm}. 

The diffusion process begins with the weighted similarity graph, $W_{\mathcal{X}}$, which is used to form the diffusion operator, $P_{\mathcal{X}}$ through row-normalization. The diffusion operator is used to learn the global data structure from the local similarities, simulating all possible transitions from point $x_i$ to point $x_j$ in $t$ discrete time steps, assuming the transition can be made in $t$ steps. In this sense $t$ can be viewed as a scale parameter: by increasing $t$, the aggregate local influence of nearest neighbors is reflected in the connectivity of the graph's points~\cite{lafon2006datafusion}. We automatically select $t$ using Von Neumann entropy as introduced in~\cite{moon2019phate}.

By connecting the weighted graphs from domains $\mathcal{X}$ and $\mathcal{Y}$ through anchor points and nearest-neighbor connections, we form an inter-domain diffusion operator, $\mathcal{P}(i, j) = \frac{W(i, j)}{\sum_{j}{W(i, j)}}$. This provides a scale-invariant~\cite{lafon2006datafusion} and stable means of estimating geodesic distances across the multimodal data structure. Instead of relying on a single shortest path, the cross-domain diffusion operator learns the global structure from the most probable paths from $x_i$ to $y_j$ in $t$ discrete steps. Paths with small transition probabilities (e.g., due to noise perturbations), are filtered through the diffusion process~\cite{coifman2006dm}, thus making the learned inter-domain distances more robust to noise than shortest paths using geodesic distances~\cite{lafon2006dm-course-grain}.

The shared powered diffusion operator, $\mathcal{P}^t$ encodes the inter-domain global transition structure where each row forms a set of $t + 1$ time-step inter-domain transition probabilities. To encode the integrated global data structure, we apply an information distance (e.g., the Hellinger distance, KL divergence, potential distance~\cite{moon2019phate}, or the dynamical set of distances given in~\cite{duque2019dig}) to the powered diffusion operator. We call the result the integrated diffusion distance, $\mathcal{D}^{t}$ which is then scaled using 0-1 normalization. The result can be embedded into lower dimensions via eigendecomposition, MDS, or other embedding methods, though we focus our results on MDS embeddings. 

The alignment process can be improved by iteratively integrating new ``pseudo-connections'' after the initial alignment process. To add new connections the user may provide a threshold, $\eta$, and limit the number of new connections that can be made during each iteration. New edge connections are assigned in the matrices $W_{\mathcal{X}\mathcal{Y}}$ and $W_{\mathcal{Y}\mathcal{X}}$ if the values of the integrated diffusion distance are less than $\eta$. That is,

\[W_{\mathcal{X}\mathcal{Y}}(i, j) = \nu - \mathcal{D}^t(i, j) \text{, if }  \mathcal{D}^t(i, j) < \eta \]

At each iteration, we evaluate the alignment by calculating the FOSCTTM score (See Section~\ref{sec:metrics}) of a set of held-out, known connections. If the alignment improves, we retain the resulting integrated diffusion distances. Otherwise, we revert to the original alignment, and the new connections are flagged to be excluded from consideration in subsequent iterations. The process of adding connections continues until (1) the maximum number of iterations is reached or (2) no new connections with values within the threshold are found. In our results, we compare MASH with and without pseudo connections, naming the method without pseudo-connections MASH-.

\begin{figure}
    \centering
    \includegraphics[width=0.8\linewidth]{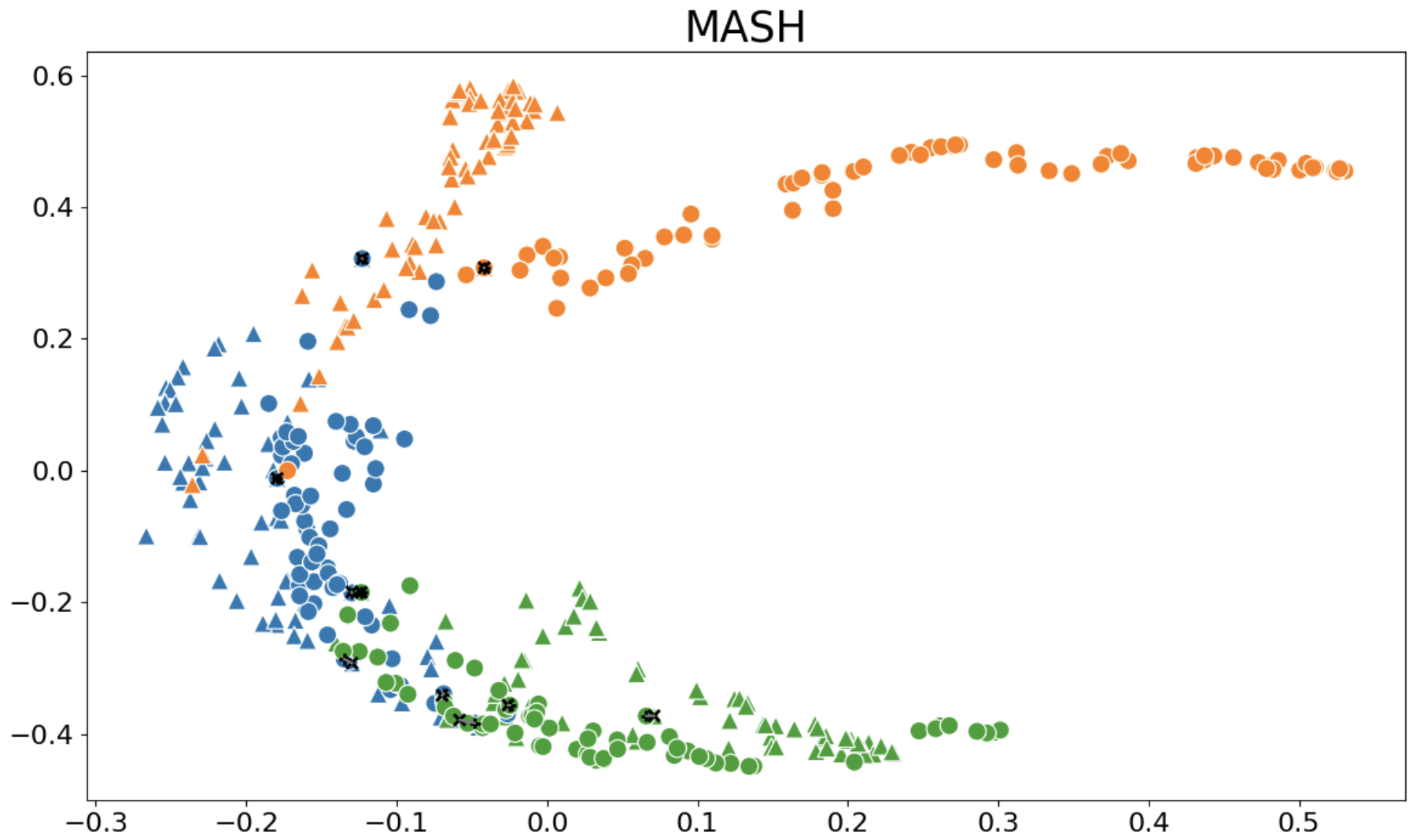}
    \caption{\footnotesize The MDS embedding of the MASH alignment of the Seeds dataset using the potential distance. The two domains are split based on the skewed split described in Section\ref{sec:splits}. The domain consisting of meaningful features is denoted by triangles and that of less relevant features by circles. The three colors denote the classes present in the dataset. Known anchor points, 5\% of the data, are colored black with gray lines connecting the corresponding anchor points. Shorter lines mean better alignment. Most of the randomly determined anchoring points belong to the blue or green classes, suggesting a reason for the divergence of the orange class branch. The embedding CE score is 88\% and its FOSCTTM score is 11\%.}
    \label{fig:MASH-example}
\end{figure}

Figure~\ref{fig:MASH-example} gives a scatterplot of an MDS embedding applied to the MASH alignment method. In the figure, we see the importance of the anchoring points in defining the alignment. Very few of the anchor points belong to the orange class. Due to this, we see a divergence in the manifold branches.  Most of the anchor points belong to the other two classes, providing better alignment of these groups. Only 5\% of these points are known correspondences.

% \subsection{Sparse Label Transfer}

\begin{figure}[!htb]
    \centering
    \includegraphics[width=0.75\linewidth]{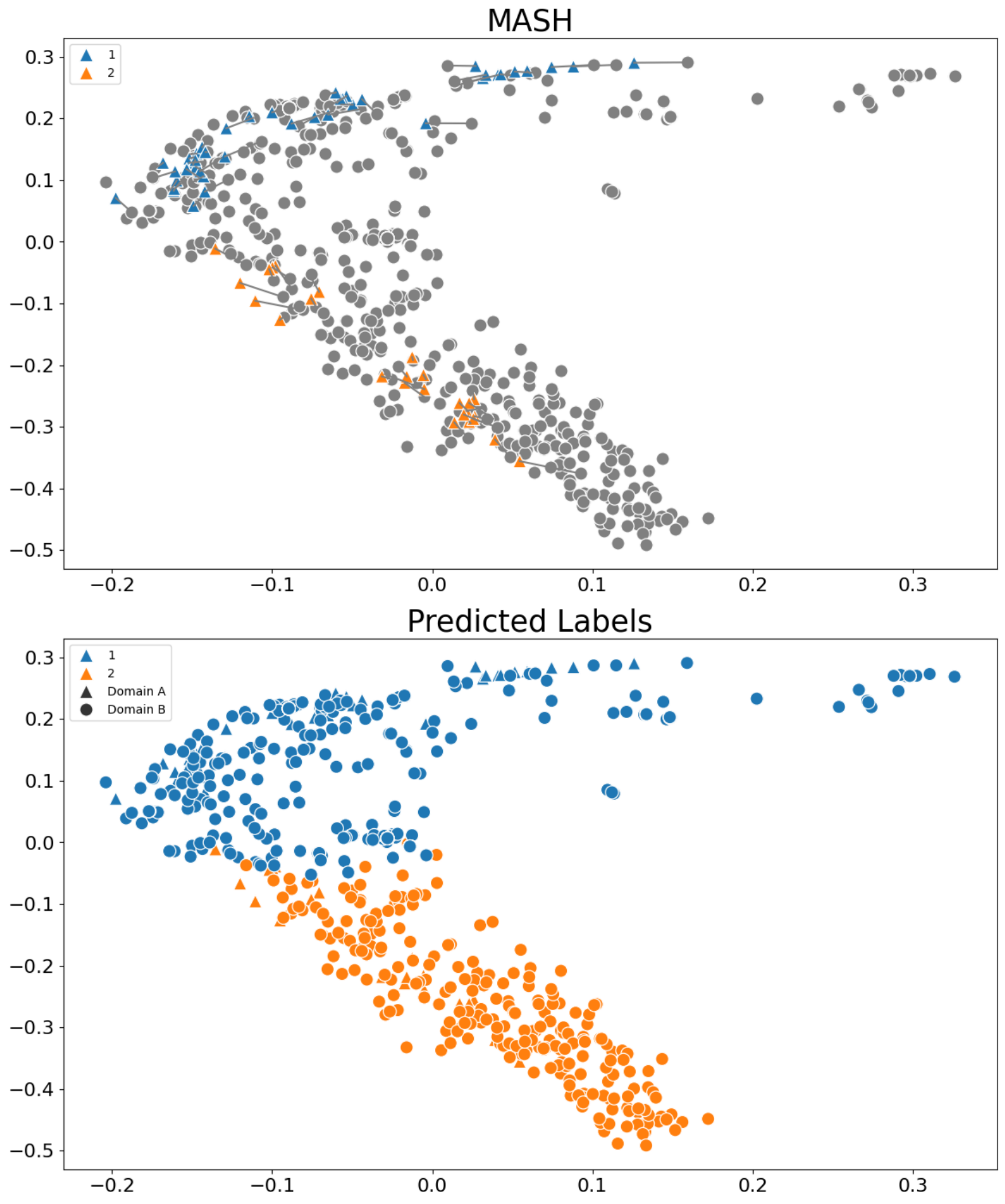}
    \caption{\footnotesize The MDS embedding of the MASH alignment of the breast cancer dataset where the 4 features with the greatest classification importance are given to Domain A, and the other five are given to Domain B. Domain A is denoted by triangles and Domain B is denoted by circles. Blue and orange denote two classes present in the dataset: whether or not the patient has breast cancer. In this example, we use the labeled information in Domain A (which has 69 data points) to predict labels across Domain B (which has 699 data points). It has an accuracy score of 97\% whereas the baseline test score is only 93\%.}
    \label{fig:Spare Label Imputation}
\end{figure}

We demonstrate MASH's ability to transfer label information between domains using the breast cancer dataset~\cite{dua2019uci}. We created domain A with the four most important features and domain B with the five remaining features. Domain A was limited to 10\% of the data observations, while domain B included all observations. There are no shared features between the domains, but 10\% of domain B's data points have alternative representations in domain A. A joint manifold with 4 dimensions was formed using MASH, and a $k$-NN model trained on the embedded representation of domain A was used to predict the labels of domain B, achieving 97\% accuracy. A 2D MDS representation is shown in Figure~\ref{fig:Spare Label Imputation}. Code for both SPUD and MASH can be found at \href{https://github.com/rustadadam/mashspud}{https://github.com/rustadadam/mashspud}.

\section{Experimental Setup}

\begin{figure*}[!htb]
    \centering
    \includegraphics[width=0.62\linewidth]{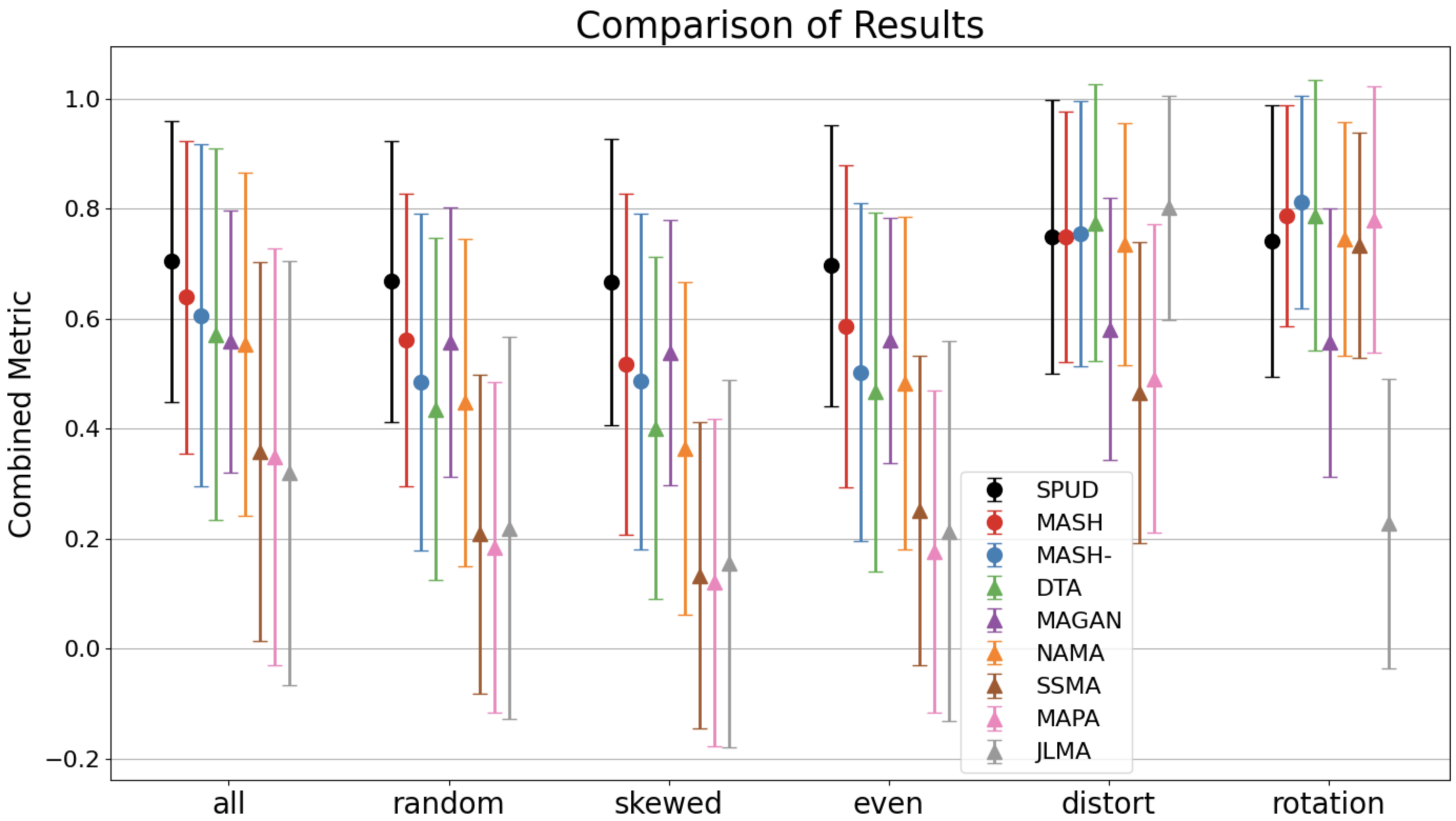}
    \caption{\footnotesize Each of the methods is compared across each split type. Our methods are denoted by dots. The methods are scored according to their average combined score (CE - FOSCTTM). The results are aggregated across all 29 datasets and 10 repetitions. A grid of parameters was searched for each method and the best set of parameters per method, dataset, split, and random state was used and recorded, testing each dataset with a randomized selection of 20\% of the points as anchors. Each feature-based split is dominated by SPUD, followed by MASH, MASH-, and MAGAN. DTA outperforms other methods at the distortion adaptation followed by JLMA and our methods. MASH- is the best-performing method for the rotated data, followed by DTA, MAPA, and MASH.}
    \label{fig:Comparison of Results}
\end{figure*}

Though sources for multimodal data are abundant, amassing datasets collected from multiple sources in a user-friendly format for large-scale experimentation is challenging. To source multimodal datasets to compare methods, we start with a collection of publicly available datasets and apply several domain adaptations in the form of feature-level splits and data distortions to artificially create alternative domains. Here we provide details about the data, adaptations, and general experimental setup.
 
\subsection{Datasets}

We compared results in 29 publicly available datasets. The majority of the datasets used in our experiments can be found on the UC Irvine Machine Learning Repository~\cite{dua2019uci}. All additional sources are cited separately. To process each dataset, observations with missing values were removed and the data were normalized at the feature level. The complete list of datasets can be found in the footnote\footnote{Artificial Tree~\cite{moon2019phate}, Audiology, Balance Scale, Breast Cancer, Car Evaluation, Chess (King-Rook vs. King-Pawn), Diabetes, Ecoli (5 majority classes), Flare1, Glass, Heart Disease, Heart Failure, Hepatitis, Hill Valley, Ionosphere, Iris, Medical Dataset~\cite{brar2024medicaldataset}, Optical Digits, Parkinson's, Seeds, Segmentation, Tic-Tac-Toe, Titanic~\cite{harrell2017titanic}, Tree Structure~\cite{moon2019phate}, Water Potability~\cite{ck2024waterprobability}, Waveform, Wine Quality, Zoo.}. We selected these datasets on the basis of their availability and minimal pre-processing requirements. Each dataset comes with categorical labels, which are not used in the comparative methods but provide a means to evaluate the cross-embedding classification performance of each method.

\subsection{Domain Adaptations}\label{sec:splits}

We performed various domain adaptations to form multiple domains from each dataset. The following adaptations are made at the feature level. That is, we subset the data according to different feature criteria. The parenthetical terms show the adaptation name which we use in Figure~\ref{fig:Comparison of Results}.

\begin{enumerate}
    \item \textit{Random (random)} -- random features were assigned to each domain. Each set of features is unique to each domain with no repeated features between domains.

    \item \textit{Skewed Importance (skewed)} -- Feature importance was determined by random feature permutations using a random forest model. The more useful half of the features form domain $\mathcal{X}$, while the less relevant half forms domain $\mathcal{Y}$.

    \item \textit{Even Importance (even)} -- Half of the features important for prediction are randomly assigned to form $\mathcal{X}$, and the others to $\mathcal{Y}$. The remaining less relevant features are randomly distributed to each domain. 

\end{enumerate}   

In addition to feature-level splits, other domain adaptations take the form of data distortion. These types of data distortion were used to compare alignment methods in~\cite{duque2023dta}.

\begin{enumerate}
    \item \textit{Gaussian Noise (distort)} -- Additive Gaussian noise is assigned to each feature of a dataset. The noisy data is designated as domain $\mathcal{Y}$. For our experiments, we set the scale factor to 0.05.

    \item \textit{Random Rotation (rotation)} -- Random rotations were applied to $\mathcal{X}$ to form $\mathcal{Y}$. The random rotations were determined using QR factorization on a randomly generated matrix. This differs from the rotations given in~\cite{duque2023dta} which randomly rotated images.
\end{enumerate}

For each domain adaptation method, anchor points were randomly assigned. We assigned known correspondence at levels of 5\%, 10\%, 15\%, 20\%, 30\%, and 50\%. Each adaptation and correspondence assignment was repeated 10 times, using consistent random seeds to validate each method.

\subsection{Compared Methods}

We compare MASH and SPUD with semi-supervised alignment methods with code or pseudocode provided by the authors, including DTA~\cite{duque2023dta}, MAGAN~\cite{amodio2018magan}, JLMA~\cite{wang2011manifold}, MAPA~\cite{wang2008ma-procrustes}, and SSMA~\cite{ham2005ssma}. Additionally, we include an early implementation of SPUD, which uses all pairwise distances within each split rather than the $k$-NN distances only to form the shortest paths and only calculates the minimum geodesic distance as an aggregation function. We call this unpublished method Nearest Anchor Manifold Alignment or NAMA.

\subsection{Metrics for Comparisons}\label{sec:metrics}

% Better describe FOSCTTM, CE, and any others
We use two metrics to quantitatively compare methods. First, we calculate the average fraction of samples closer than the true match, or FOSCTTM, which has been used for comparison in other manifold alignment papers~\cite{duque2023mali, demetci2022scot, liu2019mmd-ma}. Lower scores indicate better alignment; indicating corresponding points are mapped near each other through the alignment process. If a method perfectly aligns all true correspondences, the average FOSCTTM score would be 0. This metric does not require any label information and measures the degree of correspondence at the observation level. This metric requires points to have a true match and thus cannot be used for partial or unsupervised alignment. 

Second, we determine the accuracy of label transfer from domain $\mathcal{X}$ to domain $\mathcal{Y}$. To do so, we form a joint embedding using multidimensional scaling and train a $k$-NN model using the embedded points from $\mathcal{X}$ to predict the labels of the points in $\mathcal{Y}$. We call this method \textbf{c}ross-\textbf{e}mbedding classification or CE. This method determines the extent to which important information for labeling in one domain is transferred to the aligned embedding. For each dataset and adaptation, we set the number of embedding components to the number of features in the smaller split. We note that this quantitative method is only suitable for datasets with categorical labels, which is not a requirement for our alignment methods.

Higher CE scores mean a better alignment regarding supervision, while lower FOSCTTM scores mean a better point-wise alignment. We form a combined metric by subtracting the FOSCTTM scores from the CE scores. Thus, values of the combined metric closer to 1 are ideal for alignment.

\section{Results}

The aggregated results from the combined CE - FOSCTTM analysis, comparing all methods across each split and dataset, are shown in Figure~\ref{fig:Comparison of Results}. In the figure, we randomly selected 20\% of points to serve as anchors. Similar patterns emerge for other percentages. SPUD consistently outperforms all other methods in the feature-level splits, with MASH  and MASH- (MASH without pseudo-connections) following respectively. For the random rotation split, diffusion-based methods and MAPA predominate, with MASH- providing the best alignment, closely followed by DTA, MAPA, and MASH. DTA achieves the best results on the distortion split, closely followed by JLMA, MASH-, and MASH. 

\begin{figure}[!htb]
    \centering
    \begin{subfigure}[t]{\linewidth}
        \centering
        \includegraphics[width=0.86\linewidth]{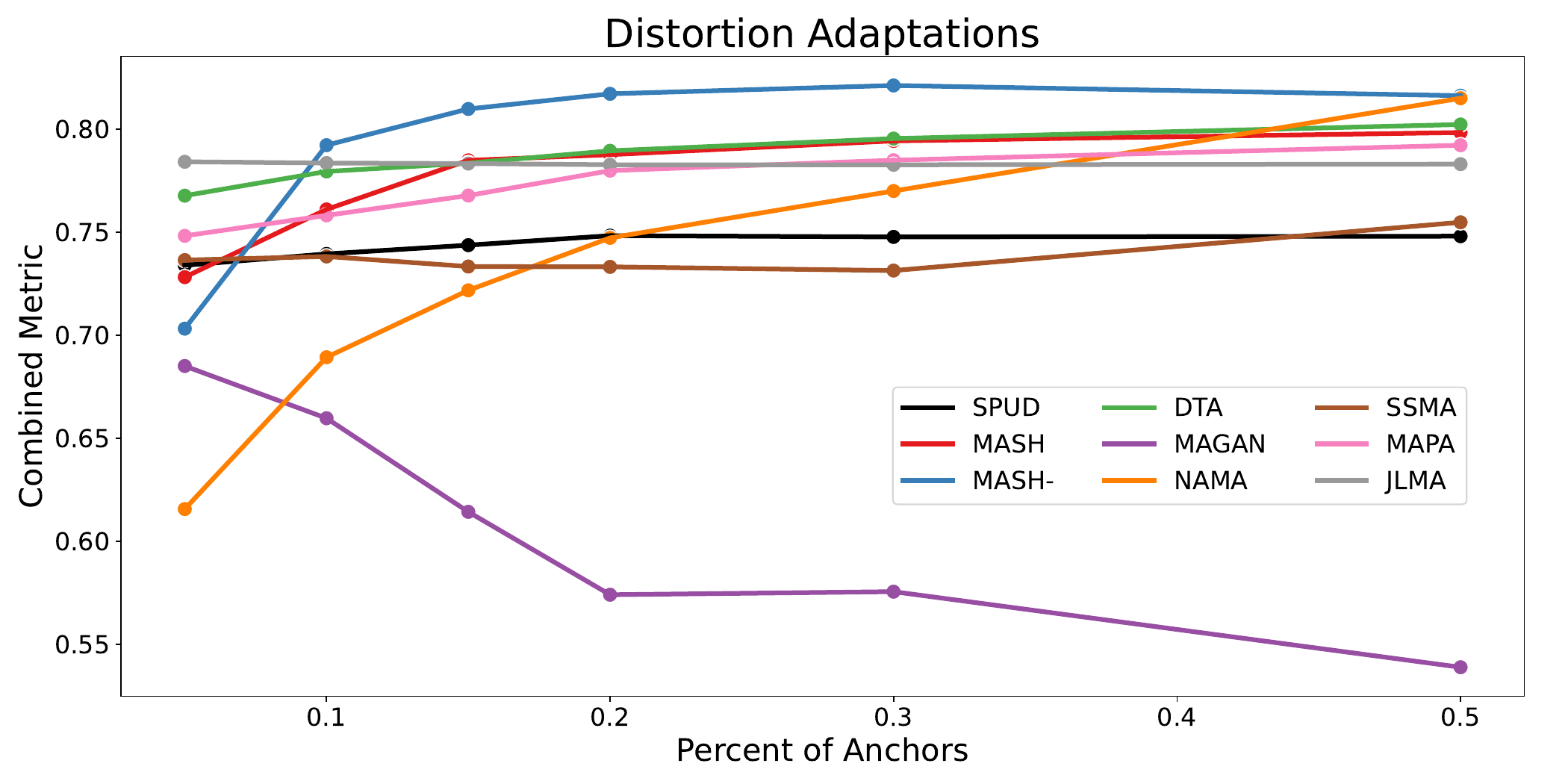}
        % \caption{}
        \label{fig:distortion-adaptations}
    \end{subfigure}
    \vspace{1em}
    \begin{subfigure}[t]{\linewidth}
        \centering
        \includegraphics[width=0.85\linewidth]{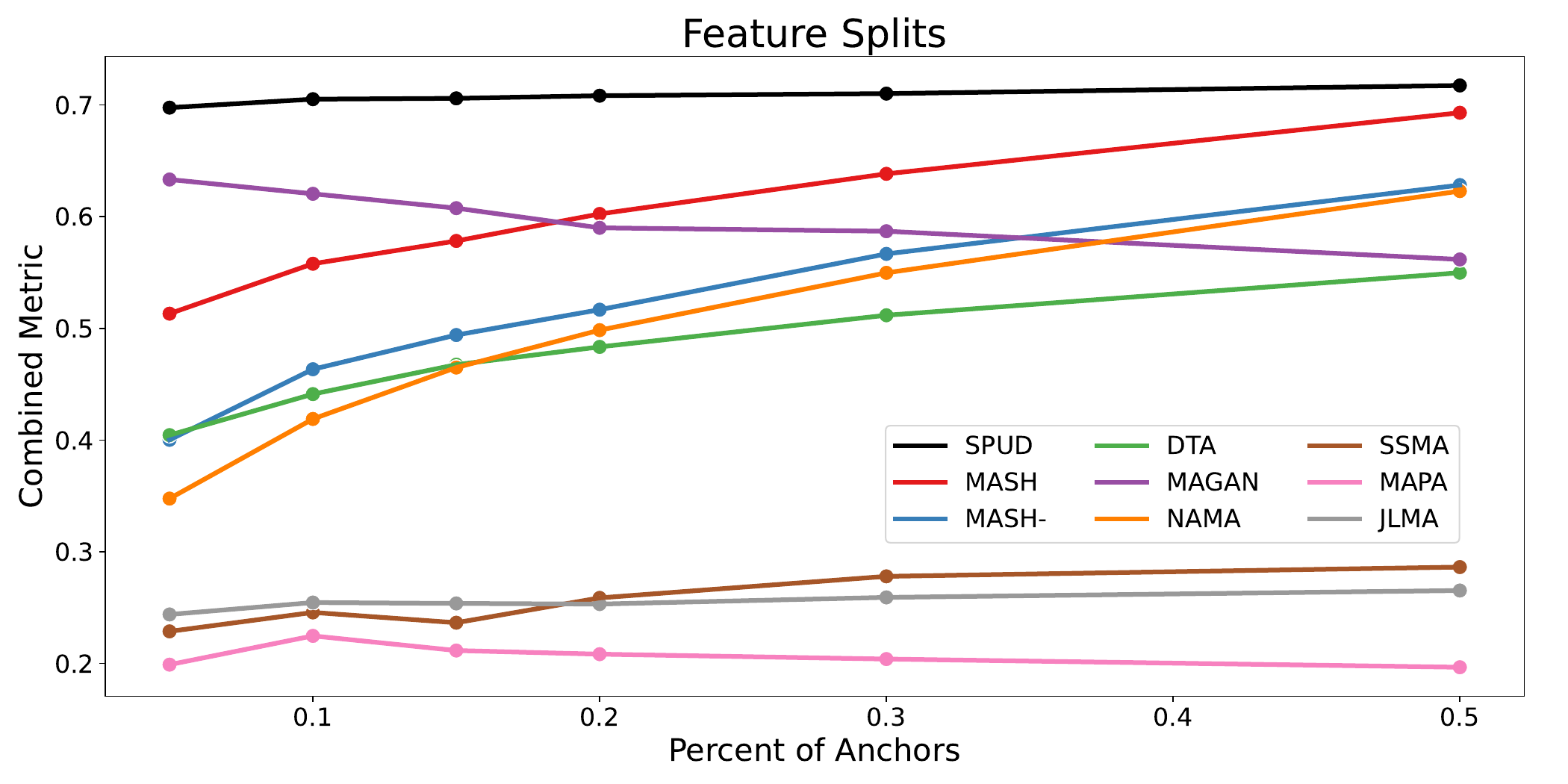}
        % \caption{TEXT HERE}
        \label{fig:feature-splits}
    \end{subfigure}
    \caption{\footnotesize Here we present the combined metric score aggregated at each percentage of known correspondence. We split the figure into two parts: (Top) feature level splits and (Bottom) distortion adaptations. The feature-level splits are dominated by SPUD across all levels. MASH- outperforms other methods at all levels at or above 10\% on distorted data. Our methods are more reliant on known correspondence than other methods at the distortion splits but do comparably well or better given at least 10\% of correspondences are known. We note that MAGAN is likely overfitting the correspondence (similar to mode collapse), leading to worse alignments at higher percentages.}
    \label{fig:combined-splits}
\end{figure}

We demonstrate comparative results at different percentages in Figure~\ref{fig:combined-splits}. The upper subfigure shows the alignment trends of each method using the feature-level splits. At each level, SPUD outperforms all other methods, only marginally improving as the number of anchors increases. MASH outperforms the other compared methods when using at least 20\% of known correspondences. The bottom subfigure compares methods using the distortion adaptations. MASH and MASH- seem to rely more on anchoring points for better performance. MASH- outperforms all other methods given 10\% or more correspondence, but has less alignment capabilities at a 5\% level. DTA and MASH have comparable performance at most percentages, but DTA has a stronger alignment at only the 5\% correspondence.

% -------------------Vertical Layout--------------------------------%

\section{Conclusion}

In this paper, we present two novel semi-supervised manifold alignment methods that leverage partial correspondences between datasets to learn cross-domain similarities. The first method, SPUD, can be regarded as an extension of Isomap to multiple domains. It initializes the alignment process by learning cross-domain geodesic distances, effectively capturing the intrinsic geometry of the datasets across different domains. The second method, MASH, is inspired by Diffusion Maps. It constructs a multimodal diffusion operator that is used to discover new, previously unknown connections between points in different domains. Both SPUD and MASH generate low-dimensional data representations by applying MDS to the learned multimodal distances. Our experimental results demonstrate that SPUD and MASH outperform or perform comparably to existing semi-supervised manifold alignment methods in two key aspects: (1) preserving distance between points of linked domains and (2) effectively transferring important features for supervised learning into the aligned multimodal manifold embedding. 

% \section*{Acknowledgement}

% The work presented in this article was in part funded by Simmons Research Endowment.

% \clearpage % I added this

\bibliographystyle{IEEEtran}
% \bibliography{references}
\bibliography{IEEEabrv, references} 

\end{document}